\documentclass[11pt,a4paper]{article}
\usepackage{times,latexsym}
\usepackage{url}
\usepackage[T1]{fontenc}

\usepackage{graphicx}
\usepackage{amssymb}
\usepackage{amsmath}
\usepackage{amsfonts}
\usepackage{mathtools}
\usepackage{booktabs}
\usepackage{graphicx}
\usepackage{xcolor}
\usepackage{enumitem}

\DeclareMathOperator{\attention}{Attention}

\DeclareMathOperator{\lstm}{LSTM}
\DeclareMathOperator{\dropout}{BDrop}
\DeclareMathOperator{\affine}{Affine}
\DeclareMathOperator{\softmaxout}{SoftmaxOut}
\usepackage[acceptedWithA]{tacl2018v2}

\usepackage{xspace,mfirstuc,tabulary}

\newif\iftaclinstructions
\taclinstructionsfalse 
\iftaclinstructions
\renewcommand{\confidential}{}
\renewcommand{\anonsubtext}{(No author info supplied here, for consistency with
TACL-submission anonymization requirements)}
\newcommand{\instr}
\fi

\iftaclpubformat 

\else

\fi

\title{Attention-Passing Models for \\Robust and Data-Efficient End-to-End Speech Translation}

       \author{Matthias Sperber$^1$, Graham Neubig$^2$, Jan Niehues$^1$, Alex Waibel$^{1,2}$ 
       		\\ $^1$Karlsruhe Institute of Technology, Germany
		\\ $^2$Carnegie Mellon University, USA
		\\ \texttt{\{first\}.\{last\}@kit.edu, gneubig@cs.cmu.edu}
		}
\date{}

\begin{document}
\maketitle
\begin{abstract}
Speech translation has traditionally been approached through cascaded models consisting of a speech recognizer trained on a corpus of transcribed speech, and a machine translation system trained on parallel texts. Several recent works have shown the feasibility of collapsing the cascade into a single, direct model that can be trained in an end-to-end fashion on a corpus of translated speech. However, experiments are inconclusive on whether the cascade or the direct model is stronger, and have only been conducted under the unrealistic assumption that both are trained on equal amounts of data, ignoring other available speech recognition and machine translation corpora.

In this paper, we demonstrate that direct speech translation models require more data to perform well than cascaded models, and while they allow including auxiliary data through multi-task training, they are poor at exploiting such data, putting them at a severe disadvantage. As a remedy, we propose the use of end-to-end trainable models with two attention mechanisms, the first establishing source speech to source text alignments, the second modeling source to target text alignment. We show that such models naturally decompose into multi-task-trainable recognition and translation tasks and propose an {\it attention-passing} technique that alleviates error propagation issues in a previous formulation of a model with two attention stages. Our proposed model outperforms all examined baselines and is able to exploit auxiliary training data much more effectively than direct attentional models.
\end{abstract}

\section{Introduction}

Speech translation takes audio signals of speech as input and produces text translations as output. While traditionally realized by cascading an automatic speech recognition (ASR) and a machine translation (MT) component, recent work showed that it is feasible to employ a single sequence-to-sequence model instead \cite{Duong2016,Weiss2017,Berard2018,Anastasopoulos2018}. An appealing property of such direct models is that we no longer suffer from propagation of errors, where the speech recognizer passes an erroneous source text to the machine translation component, potentially leading to compounding follow-up errors. Another advantage is the ability to train all model parameters jointly.

\begin{figure}[t]
\includegraphics[width=8cm]{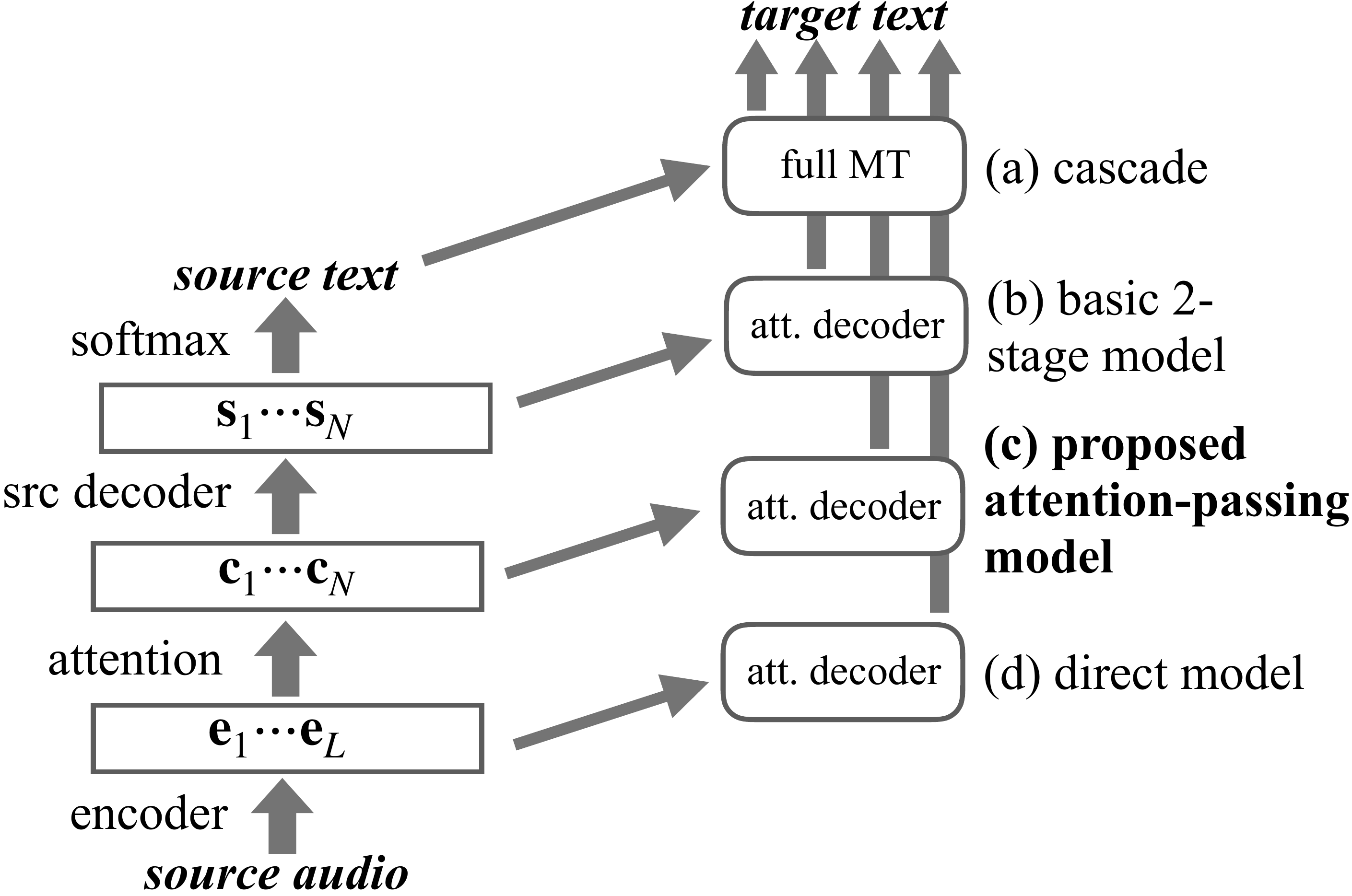}
\caption{Conceptual diagrams for various speech translation approaches. {\it Cascade} (a) uses separate machine translation and speech recognition models. The {\it direct} model (d) is a standard attentional encoder-decoder model. The {basic 2-stage} model (b) uses two attention stages and passes source-text decoder states to the translation component. Our proposed {\it attention-passing} model (c) applies two attention stages, but passes context vectors to the translation component for improved robustness.}
\label{fig:e2e-conceptual}
\end{figure}

Despite these obvious advantages, two problems persist: (1) Reports on whether direct models outperform cascaded models (Fig.~\ref{fig:e2e-conceptual}a,d) are inconclusive, with some work in favor of direct models \cite{Weiss2017}, some work in favor of cascaded models \cite{Kano2017,Berard2018}, and one work in favor of direct models for two out of the three examined language pairs \cite{Anastasopoulos2018}. (2) Cascaded and direct models have been compared under identical data situations, but this is an unrealistic assumption: In practice, cascaded models can be trained on much more abundant independent ASR and MT corpora, while end-to-end models require hard-to-acquire end-to-end corpora of speech utterances paired with textual translations.

Our first contribution is a closer investigation of these two issues. Regarding the question of whether direct models or cascaded models are generally stronger, we hypothesize that direct models require more data to work well, due to the more complex mapping between inputs (source speech) and outputs (target text). This would imply that direct models outperform cascades when enough data is available, but under-perform in low-data scenarios. We conduct experiments and present empirical evidence in favor of this hypothesis. Next, for a more realistic comparison with regards to data conditions, we train a direct speech translation model using more auxiliary ASR and MT training data than end-to-end data. This can be implemented through multi-task training \cite{Weiss2017,Berard2018}. Our results show that the auxiliary data is beneficial only to a limited extent, and that direct multi-task models are still heavily dependent on the end-to-end data.

As our second contribution, we apply a {\it two-stage} model \cite{Tu2017,Kano2017} as an alternative solution to our problem, hoping that such models may overcome the data efficiency shortcoming of the direct model. Two-stage models consist of a first-stage attentional sequence-to-sequence model that performs speech recognition and then passes the decoder states as input to a second attentional model that performs translation (Fig.~\ref{fig:e2e-conceptual}b). This architecture is closer to cascaded translation while maintaining end-to-end trainability. Introducing supervision from the source-side transcripts midway through the model creates inductive bias that guides the complex transformation between source speech and target text through a reasonable intermediate representation closely tied to the source text. The architecture has been proposed by \newcite{Tu2017} to realize a reconstruction objective, and a similar model was also applied to speech translation \cite{Kano2017} to ease trainability, although no experiments under varying data conditions have been conducted. We hypothesize that such a model may help to address the identified data efficiency issue: Unlike multi-task training for the direct model that trains auxiliary models on additional data but then discards many of the additionally learned parameters, the two-stage model uses all parameters of sub-models in the final end-to-end model. Empirical results confirm that the two-stage model is indeed successful at improving data efficiency, but suffers from some degradation in translation accuracy under high data conditions compared to the direct model. One reason for this degradation is that this model re-introduces the problem of error propagation, because the second stage of the model depends on the decoder states of the first model stage which often contain errors.

Our third contribution, therefore, is an {\it attention-passing} variant of the two-stage model that, instead of passing on possibly erroneous decoder states from the first to the second stage, passes on only the computed attention context vectors (Fig.~\ref{fig:e2e-conceptual}c). We can view this approach as replacing the early decision on a source-side transcript by an early decision only on the {\it attention scores} needed to compute the same transcript, where the attention scores are expectedly more robust to errors in source text decoding. We explore several variants of this model and show that it outperforms both the direct model and the vanilla two-stage model, while maintaining the improved data efficiency of the latter. Through an analysis, we further observe a trade-off between sensitivity to error propagation and data efficiency.

\section{Baseline Models}
\label{sec:e2e_baselines}

This section introduces two types of end-to-end trainable models for speech translation, along with a cascaded approach, which will serve as our baselines. All models
are based on the attentional encoder-decoder architecture of \newcite{Bahdanau2014} with character-level outputs, and use the architecture described in \S\ref{sec:e2e-audio-encoder} as audio encoders. The end-to-end trainable models include a direct model and a two-stage model. Both are limited\footnote{Prior work noted that in severe low-resource situations it may actually be easier to collect speech paired with translations than transcriptions \cite{Duong2016}. However, we focus on well-resourced languages for which ASR and MT corpora exist and for which it is more realistic to obtain good speech translation accuracy.} by the fact that they can {\it only} be trained on end-to-end data which is much harder to obtain than ASR or MT data used to train traditional cascades.\footnote{As a case in point, the largest available speech translation corpora \cite{Post2013,Kocabiyikoglu2018} are an order of magnitude smaller than the largest speech recognition corpora \cite{Cieri2004,Panayotov2015} ($\sim$ 200 hours vs 2000 hours) and several orders of magnitude smaller than the largest machine translation corpora, e.g.\ those provided by the Conference on Machine Translation (WMT).} \S\ref{sec:e2e-multi-task} will introduce multi-task training as a way to overcome this limitation.

\subsection{Audio Encoder}
\label{sec:e2e-audio-encoder}
Sequence-to-sequence models can be adopted for audio inputs by directly feeding speech features (here: Mel filterbank features) instead of word embeddings as encoder inputs \cite{Chorowski2015,Chan2016}. Such an encoder transforms $M$ feature vectors $\mathbf{x}_{1:M}$ into $L$ encoded vectors $\mathbf{e}_{1:L}$, performing downsampling such that $L{<}M$. 
We use an encoder architecture that follows one of the variants described by \newcite{Zhang2017e}: we stack two blocks, each consisting of a bidirectional LSTM, a network-in-network (NiN) projection that downsamples by factor two, and batch normalization. After the second block, we add a final bidirectional LSTM layer. NiN denotes a simple linear projection applied at every time step, performing downsampling by concatenating pairs of adjacent projection inputs. Because of space constraints, we do not present detailed equations, but refer interested readers to \newcite{Zhang2017e} as well as to our provided code for details. %

\subsection{Direct Model}
\label{sec:e2e-direct-model}

The sequence-to-sequence model with audio inputs outlined above can be trained as a direct speech translation model by using speech data as input and the corresponding translations as outputs. Such a model does not rely on intermediate ASR output and is therefore not subject to error propagation. However, the transformation from source speech inputs to target text outputs is much more complex than that of an ASR or MT system taken individually, which may cause the model to require more data to perform well.

To make matters precise, given $L$ audio encoder states $\mathbf{e}_{1:L}$ computed by the audio encoder as described in \S\ref{sec:e2e-audio-encoder}, the direct model is computed as
\begin{align}
  \mathbf{s}_i&=\lstm\left(\left[W_{\text{e}}y_{i-1};\mathbf{c}_{i-1}\right],\mathbf{s}_{i-1};\theta_{\text{lstm}}\right) \label{eq:direct-s} \\
  \mathbf{c}_i&=\attention\left(\mathbf{s}_i,\mathbf{e}_{1:L};\theta_{\text{att}}\right) \label{eq:direct-c} \\
  \tilde{\mathbf{s}}_i&=\tanh\left(W_{\text{s}}\left[\mathbf{s}_i;\mathbf{c}_i\right]+\mathbf{b}_\text{s}\right) \label{eq:direct-stilde} \\
  p&\left(y_i\mid y_{<i},\mathbf{e}_{1:L}\right)=\softmaxout\left(\tilde{\mathbf{s}}_i;\theta_\text{out}\right). \label{eq:direct-out}
\end{align}

Here, $W_\ast$, $\mathbf{b}_\ast$, and $\theta_\ast$ are the trainable parameters, $y_i$ are output characters, and $\softmaxout$ denotes an affine projection followed by a softmax operation. $\mathbf{s}_i$ are decoder states with $\mathbf{s}_0$ initialized to the last encoder state, and $\mathbf{c}_i$ are attentional context vectors with $\mathbf{c}_0{=}\mathbf{0}$. In equation~\ref{eq:direct-c}, we compute $\attention\left(\cdot\right){=}\sum_{j=1}^{L}\alpha_{ij}\mathbf{e}_j$ with weights $\alpha_{ij}$ conditioned on $\mathbf{e}_j$ and $\mathbf{s}_i$, parametrized by $\theta_{\text{att}}$, and normalized via a softmax operation.

\subsection{Two-Stage Model}
\label{sec:e2e-basic-two-stage-model}

\begin{figure}[tb]
\includegraphics[width=7.5cm]{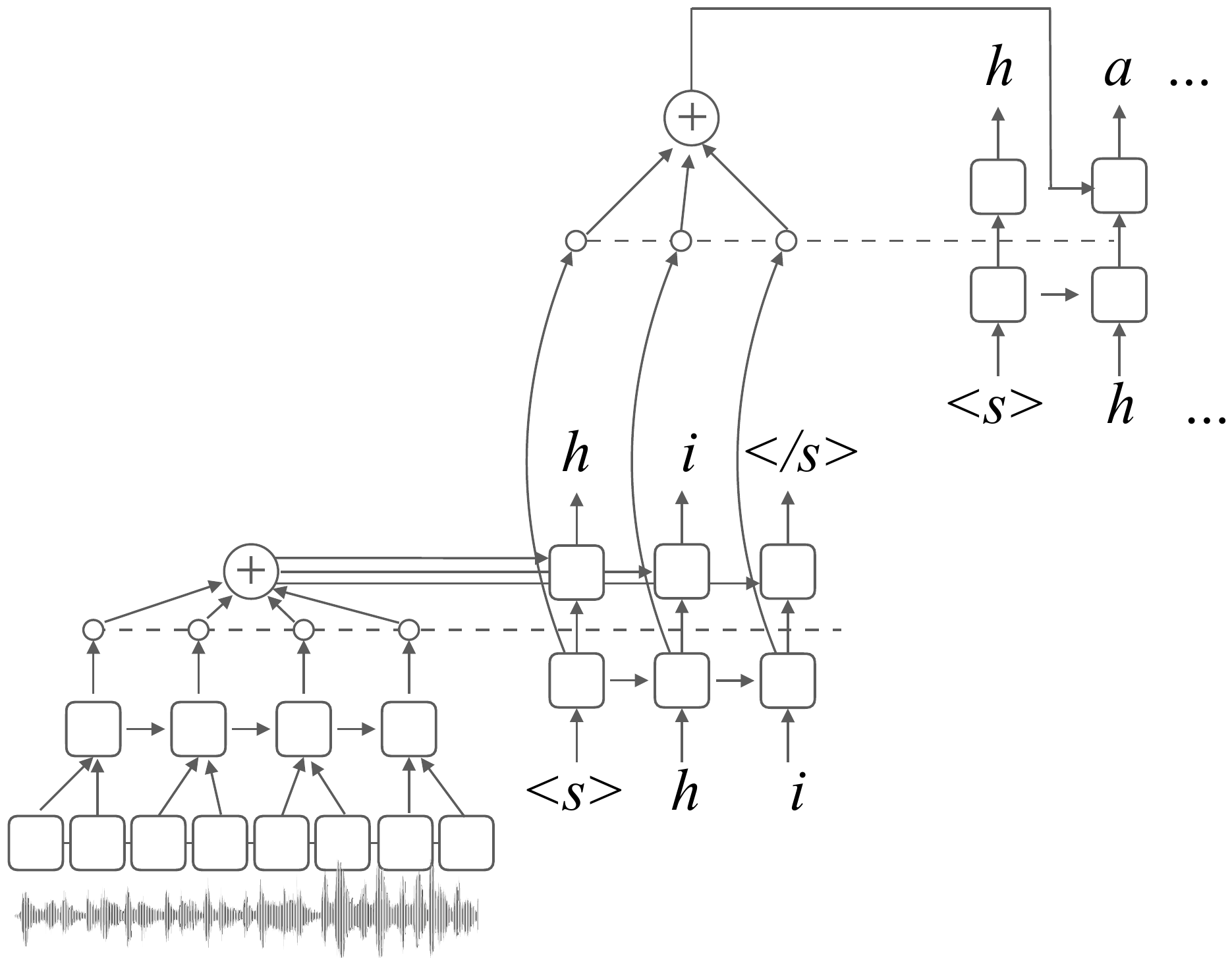}
\caption{Basic two-stage model. Decoder states of the first stage double as encoder states for the second stage.}
\label{fig:e2e-basic-multistage}
\end{figure}

As an alternative to the direct model, the two-stage model uses a cascaded information flow while maintaining end-to-end trainability. Our main motivation for using this model is the potentially improved data efficiency when adding auxiliary ASR and MT training data (\S\ref{sec:e2e-multi-task}). This model is similar to the architecture first described by \newcite{Tu2017}. It combines two encoder-decoder models in a cascade-like fashion, with the decoder of the first stage and the encoder of the second stage being shared (Fig.\ \ref{fig:e2e-basic-multistage}). In other words, while a cascade would use the source-text outputs of the first stage as inputs into the second stage, in this model the second stage directly computes attentional context vectors over the decoder states of the first stage. The inputs of the two-stage model are speech frames, the outputs of the first stage are transcribed characters in the source language, and the outputs of the second stage are translated characters in the target language. 

Again assuming $L$ audio encoder states $\mathbf{e}_{1:L}$, the first stage outputs of length $N$ are computed identically to equations \ref{eq:direct-s}--\ref{eq:direct-out}, except that input feeding (conditioning the decoding step on the previous context vector) is not used in the first stage decoder to keep components compatible for multi-task training (\S\ref{sec:e2e-multistage-multitask}):

\begin{align}
  \mathbf{s}_i^{\text{src}}&=\lstm\left(W_\text{e}^\text{src}y_{i-1}^\text{src},\mathbf{s}_{i-1}^\text{src};\theta_\text{lstm}^\text{src}\right) \label{eq:stage1-s} \\
  \mathbf{c}_i^{\text{src}}&=\attention\left(\mathbf{s}_i^\text{src},\mathbf{e}_{1:L};\theta_\text{att}^\text{src}\right) \label{eq:stage1-c} \\
  \tilde{\mathbf{s}}_i^{\text{src}}&=\tanh\left(W_\text{s}^\text{src}\left[\mathbf{s}_i^\text{src};\mathbf{c}_i^\text{src}\right]+\mathbf{b}_\text{s}^\text{src}\right) \label{eq:stage1-stilde} \\
  p&\left(y_i^\text{src}\mid y_{<i},\mathbf{e}_{1:L}\right) \nonumber\\
  &=\softmaxout\left(\tilde{\mathbf{s}}_i^\text{src};\theta_\text{out}^\text{src}\right) \label{eq:stage1-out}
\end{align}

Next, the second stage proceeds similarly but using the stage 1 decoder states as input:
\begin{align}
  \mathbf{s}_j^{\text{trg}}&=\lstm\left(\left[W_\text{e}^\text{trg}y_{i-1}^\text{trg};\mathbf{c}_{j-1}^\text{trg}\right],\mathbf{s}_{j-1}^\text{trg};\theta_\text{lstm}^\text{trg}\right) \label{eq:stage2-s} \\
  \mathbf{c}_j^{\text{trg}}&=\attention\left(\mathbf{s}_j^\text{trg},\mathbf{s}_{1:N}^\text{src};\theta_\text{att}^\text{trg}\right) \label{eq:stage2-c} \\
  \tilde{\mathbf{s}}_j^{\text{trg}}&=\tanh\left(W_\text{s}^\text{trg}\left[\mathbf{s}_j^\text{trg};\mathbf{c}_j^\text{trg}\right]+\mathbf{b}_\text{s}^\text{trg}\right) \label{eq:stage2-stilde} \\
  p&\left(y_j^\text{trg}\mid y_{<j},\mathbf{s}_{1:N}^\text{src}\right) \nonumber\\
  &=\softmaxout\left(\tilde{\mathbf{s}}_j^\text{trg};\theta_\text{out}^\text{trg}\right) \label{eq:stage2-out}
\end{align}

\subsection{Cascaded Model}
We finally employ a traditional cascaded model as a baseline, whose architecture is kept as similar to the above models as possible in order to facilitate meaningful comparisons. The cascade consists of an ASR component and an MT component, which are both attentional sequence-to-sequence models according to equations \ref{eq:direct-s}-\ref{eq:direct-out}, trained on the appropriate data. The ASR component uses the acoustic encoder of \S\ref{sec:e2e-audio-encoder}, while the MT model uses a bidirectional LSTM with 2 layers as encoder.

\section{Incorporating Auxiliary Data}
\label{sec:e2e-multi-task}
The models described in \S\ref{sec:e2e-direct-model} and \S\ref{sec:e2e-basic-two-stage-model} are trained only on speech utterances paired with translations (and transcripts in the case of \S\ref{sec:e2e-basic-two-stage-model}), which is a severe limitation.
To incorporate auxiliary ASR and MT data into the training we make use of a multi-task training strategy. Such a strategy trains auxiliary ASR and MT models that share certain parameters with the main speech translation model. We implement multi-task training by drawing several minibatches, one minibatch for each task, and performing an update based on the accumulated gradients across tasks. Note that this results in a balanced contribution of each task.\footnote{We also experimented with a final fine-tuning phase on only the main task \cite{Niehues2017}, but discarded this strategy for lack of consistent gains.}

\subsection{Multi-Task Training for the Direct Model}
\label{sec:e2e-direct-multi-task}
\begin{figure}[tb]
\includegraphics[width=7.5cm]{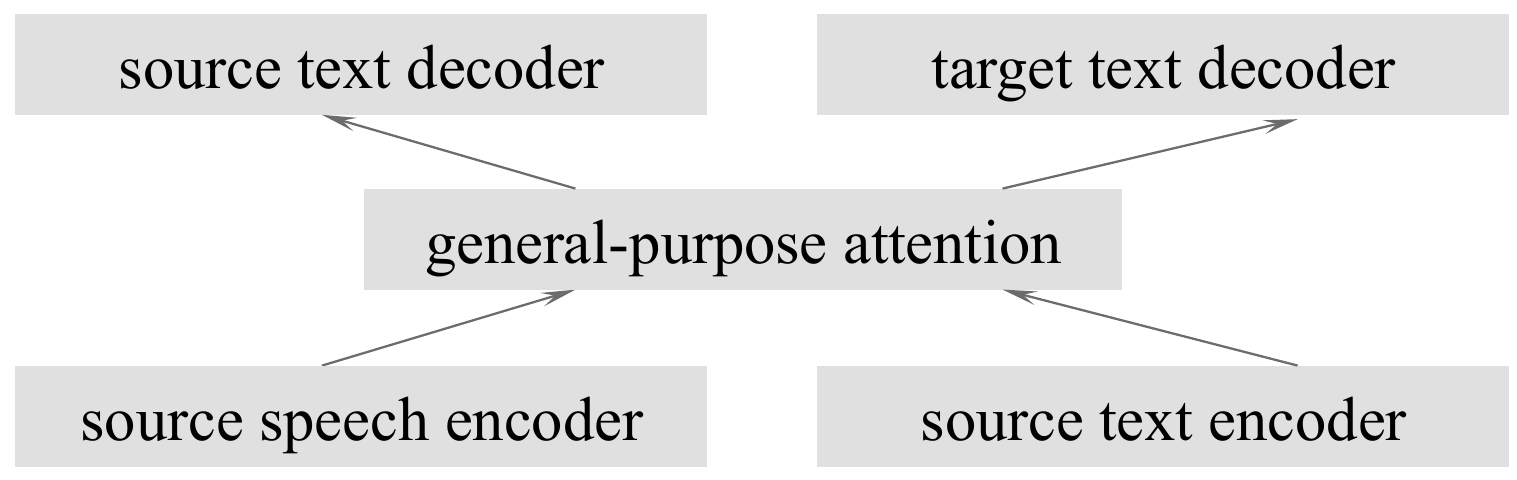}
\caption{Direct multi-task model.}
\label{fig:e2e-sharing}
\end{figure}

Multi-task training for direct speech translation models has previously been used by \newcite{Weiss2017,Berard2018}, although not for the purpose of adding additional training utterances that are not shown to the main speech translation task.\footnote{Note that \newcite{Bansal2018a} do experiment with additional speech recognition data, although, differently from our work, for purposes of cross-lingual transfer learning.} We distinguish five model components: a source speech encoder, a source text encoder (a two-layer bidirectional LSTM working on character level), a source text decoder, a target text decoder, and an attention mechanism which we opt to share across all tasks. There are four ways in which these components can be combined into a complete sequence-to-sequence model (see Figure~\ref{fig:e2e-sharing}), corresponding to the following four tasks:
\begin{description}
  \item[ASR:] Combines source speech encoder, general-purpose attention, source text decoder. This is similar to the auxiliary ASR task used by \newcite{Weiss2017} and can be trained on common ASR data.
  \item[MT:] Combines source text encoder, general-purpose-attention, target text decoder. The addition of an MT task has been mentioned by \newcite{Berard2018} and allows training on common MT data.
  \item[ST:] Combines source speech encoder, general-purpose-attention, target text decoder. This is our main task and requires end-to-end data for training.
  \item[Auto-encoder (AE):] Combines source text encoder, general-purpose attention, source text decoder. The AE task can be trained on monolingual corpora in the source language and may serve to tighten the coupling between components and potentially improves the parameters of the general-purpose attention model. We have observed slight improvements by adding the AE task in preliminary experiments and will therefore use it throughout this paper.
\end{description}

\subsection{Multi-Task Training for the Two-Stage Model}
\label{sec:e2e-multistage-multitask}
Including an auxiliary ASR task is straight-forward with the two-stage model by simply computing the cross-entropy loss with respect to the softmax output of the first stage, and dropping the second stage.

The auxiliary MT task computes only the second stage, replacing the inputs $\mathbf{s}_{1:N}^\text{src}$ by states $\mathbf{e}_{1:N}^\text{asr}$ computed as:

\begin{align}
  \mathbf{e}_{i}^\text{asr}=\lstm\left(W_\text{e}^\text{src}y_{i}^\text{transcr},\mathbf{e}_{i-1}^\text{src};\theta_\text{lstm}^\text{src}\right).
\end{align}

That is, instead of computing the second-stage inputs using the first stage, we compute these inputs through a conventional encoder that encodes the reference transcript $y_{1:N}^\text{transcr}$ and uses the same embeddings matrix and unidirectional LSTM as the first stage decoder.
Note that there is no equivalent to the auxiliary auto-encoder task of the direct multi-task model here.

Why might this architecture help to make better use of auxiliary ASR and MT data? Note that in the direct model only roughly half of the model parameters are shared between the main task and the ASR task, and likewise for main and MT tasks (\S\ref{sec:e2e-direct-multi-task}). Additional data would therefore only have a rather indirect impact on the main task. In contrast, in the two-stage model all parameters of the auxiliary tasks are shared with the main task and therefore have a more direct impact, potentially leading to better data efficiency.

Note that somewhat related to our multi-task strategy, \newcite{Kano2017} have decomposed their two-stage model in a similar way to perform pretraining for the individual stages, although not with the goal of incorporating additional auxiliary data.

\section{Attention-Passing Model}
\label{sec:e2e-modified-model}

We have so far described a direct model that has the appealing property of avoiding error propagation in a principled way but that may not be particularly data efficient, and have described a two-stage model that addresses the latter disadvantage. Unfortunately, the two-stage model re-introduces the error propagation problem back into end-to-end modeling, because the second stage heavily depends on the potentially erroneous decoder states of the first stage. We therefore propose an improved {\it attention-passing} model in this section that is less impacted by error propagation issues.

\subsection{Preventing Error Propagation}
\label{sec:e2e-modified-model-basic}
\begin{figure}[tb]
\includegraphics[width=7.5cm]{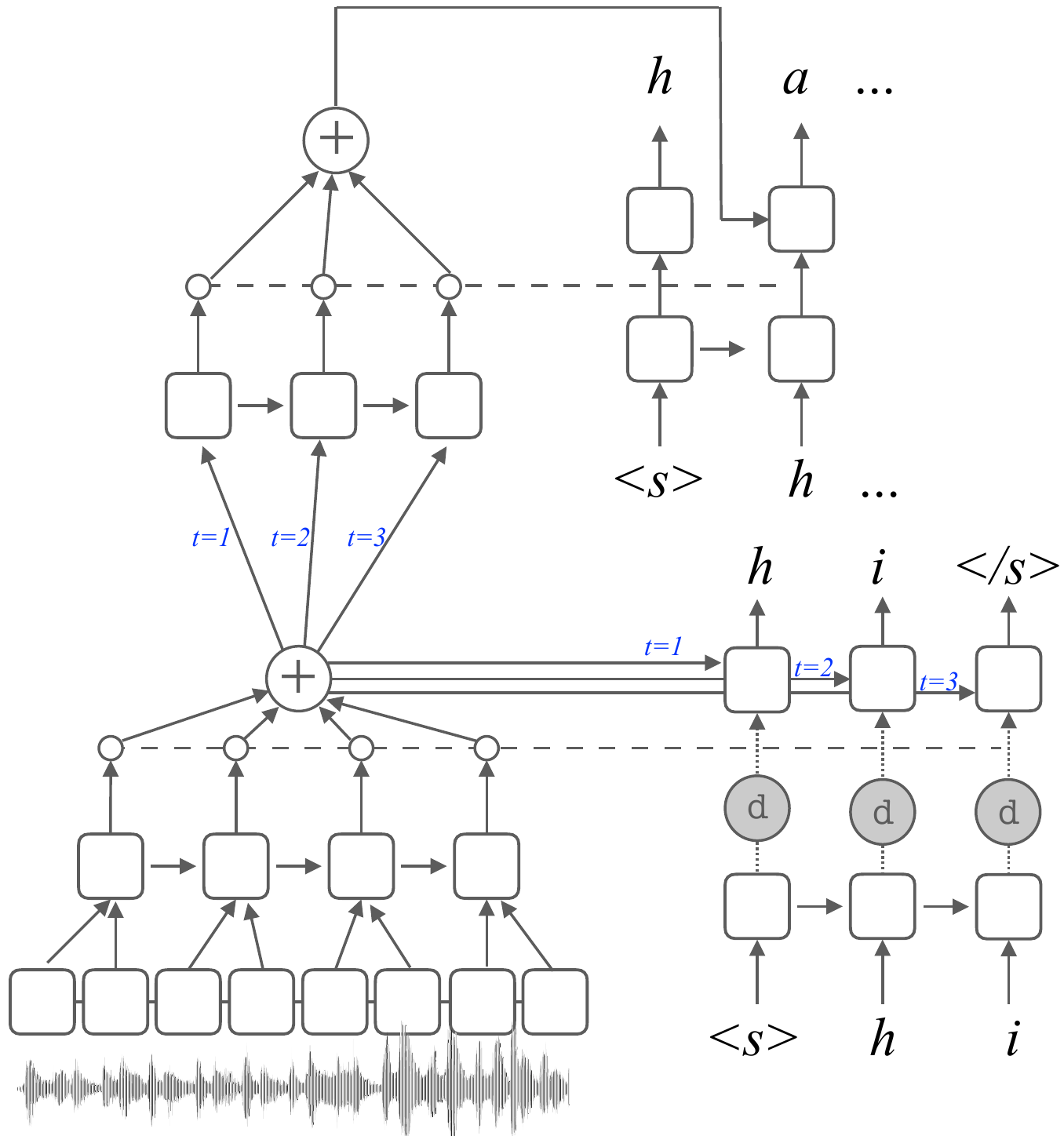}
\caption{Attention-passing model.}
\label{fig:e2e-improved-multistage}
\end{figure}

The main idea behind the attention-passing model is to not feed the erroneous first-stage decoder states to the second stage, but instead to pass on only the {\it context vectors} that summarize the relevant encoded audio at each decoding step. The first stage decoder is unfolded as usual by employing discrete source-text representations, but the only information exposed to the translation stage are the per-timestep context vectors created as a by-product of the decoding. Figure~\ref{fig:e2e-improved-multistage} illustrates this idea. Intuitively, we expect this to help because we no longer make an early decision on the identity of the source-language text, but only on the corresponding attentions. This is motivated by our observation that speech recognition attentions are sufficiently robust against decoding errors (\S\ref{sec:e2e-robust_attentions}).

Formally, the first stage remains unchanged from equations \ref{eq:stage1-s}--\ref{eq:stage1-out}. The context vectors $\mathbf{c}_i^\text{src}$ then form the input to the second stage:

\begin{align}
  \mathbf{x}_i^{\text{trg}}&=\lstm\left(\mathbf{c}_i^\text{src},\mathbf{x}_{i-1}^\text{trg};\theta_\text{lstm}^\text{src}\right) \label{eq:improved2-x} \\
  \mathbf{s}_j^{\text{trg}}&=\lstm\left(\left[W_\text{e}^\text{trg}y_{i-1}^\text{trg};\mathbf{c}_{j-1}^\text{trg}\right],\mathbf{s}_{j-1}^\text{trg};\theta_\text{lstm}^\text{trg}\right) \label{eq:improved2-s} \\
  \mathbf{c}_j^{\text{trg}}&=\attention\left(\mathbf{s}_j^\text{trg},\mathbf{x}_{1:N}^\text{trg};\theta_\text{att}^\text{trg}\right) \label{eq:improved2-c} \\
  \tilde{\mathbf{s}}_j^{\text{trg}}&=\tanh\left(W_\text{s}^\text{trg}\left[\mathbf{s}_j^\text{trg};\mathbf{c}_j^\text{src}\right]+\mathbf{b}_\text{s}^\text{trg}\right) \label{eq:improved2-stilde} \\
  p&\left(y_j^\text{trg}\mid y_{<j},\mathbf{s}_{1:N}^\text{src}\right)\nonumber\\
  &=\softmaxout\left(\tilde{\mathbf{s}}_j^\text{trg};\theta_\text{out}^\text{trg}\right) \label{eq:improved2-out}
\end{align}

\subsection{Decoder State Drop-Out}

Along with the modifications described in \S\ref{sec:e2e-modified-model-basic}, we introduce an additional block drop-out operation \cite{Ammar2016} on the decoder states, replacing equation~\ref{eq:stage1-stilde} by 
$$
\tilde{\mathbf{s}}_i^{\text{src}}=\tanh\left(W_\text{s}^\text{src}\left[\dropout\left\{\mathbf{s}_i^\text{src}\right\};\mathbf{c}_i^\text{src}\right]+\mathbf{b}_\text{s}^\text{src}\right). \label{eq:improved1-stilde}
$$
The block drop-out operation, denoted as $\dropout$, replaces the whole vector by zero with a certain probability (here: $0.5$). This results in the context vectors $\mathbf{c}_i^\text{src}$ becoming the only  information available to the output layer whenever the decoder states are dropped out. The motivation for this is to force the model to maximize the informativeness of the context vectors, which are later relied upon as sole inputs to the second stage.

\subsection{Multi-Task Training}
Similar to the basic two-stage model, the attention-passing model as a whole is trained on speech-transcript-translation triplets, but can be decomposed into two sub-models that correspond to ASR and MT tasks. In fact, the ASR task is unchanged with the exception of the new block dropout operation. The MT task is obtained by replacing equation~\ref{eq:improved2-x} by $\mathbf{x}_i^{\text{trg}}=\lstm\left(W_{e}\mathbf{\mathrm{y}}_{i}^\text{src},\mathbf{x}_{i-1}^\text{trg};\theta_\text{lstm}^\text{src}\right)$, i.e.\ by using the transcript character embeddings as inputs instead of the context vectors used when training the main task. Note that the LSTMs in equations \ref{eq:stage1-s} and \ref{eq:improved2-x} are shared in order to have a match between stage 1 decoder and stage 2 encoder as with the basic model.

\subsection{Cross Connections}
\label{sec:e2e-cross-connections}

As a further extension to the attention-passing model of \S\ref{sec:e2e-modified-model-basic}, we can introduce cross connections that concatenate the dropped-out first stage hidden decoder states to the second-stage inputs encoder. This causes equation~\ref{eq:improved2-x} to be replaced by 

\begin{align}
\mathbf{x}_i^{\text{trg}}=&\\
  \lstm&\left(\affine\left[\mathbf{c}_i^\text{src};\dropout\left\{\mathbf{s}_i^\text{src}\right\}\right],\mathbf{x}_{i-1}^\text{trg};\theta_\text{lstm}^\text{src}\right) \nonumber\label{eq:improved2-x-cross}
\end{align}

This extension moves the model closer to the basic two-stage model, while the inclusion of the context vectors and the block drop-out operation on the hidden decoder states ensures that the second stage decoder does not rely too strongly on the first stage outputs. 

\subsection{Additional Loss}
\label{sec:e2e-add-loss}
We further experiment with introducing an additional loss aimed at making the LSTM inputs between first stage decoder and second stage encoder RNN more similar. Recall that in our attention-passing model, both RNNs share parameters (equations \ref{eq:stage1-s} and \ref{eq:improved2-x}), so that similar inputs at both times is desirable. The loss is defined as follows: $$\mathcal{L}_\text{add}=||\mathbf{c}_{i}^\text{src}-W_{e}\mathbf{\mathrm{y}}_{i}^\text{src}||_{2}.$$

If combined with the cross connections (\S\ref{sec:e2e-cross-connections}), the formula is adjusted to $\mathcal{L}_\text{add}=||\affine\left[\mathbf{c}_i^\text{src};\dropout\left\{\mathbf{s}_i^\text{src}\right\}\right]-W_{e}\mathbf{\mathrm{y}}_{i}^\text{src}||_{2}$.
We did not find it beneficial to apply a scaling factor when adding this loss to the main cross-entropy loss in our experiments.

\section{Experiments}
\label{sec:experiments}

We conduct experiments on the Fisher and Callhome Spanish--English Speech Translation Corpus \cite{Post2013}, a corpus of Spanish telephone conversations that includes audio, transcriptions, and translations into English. We use the Fisher portion that consists of telephone conversations between strangers. The training data size is 138,819 sentences, corresponding to 162 hours of speech. ASR word error rates (WER) on this dataset are usually relatively high due to the spontaneous speaking style and challenging acoustics. From a translation viewpoint, the data can be considered as relatively easy with regards to both the topical domain and particular language pair.

Our implementation is based on the \texttt{xnmt} toolkit.\footnote{\url{https://github.com/neulab/xnmt}} We use the speech recognition recipe as a starting point, which has previously been shown to achieve competitive ASR results \cite{neubig18xnmt}. Code and configuration files can be found at \url{http://www.msperber.com/research/tacl-attention-passing/}.

The vocabulary consists of the common characters appearing in English and Spanish, apostrophe, whitespace, and special start-of-sequence and unknown-character tokens. The same vocabulary is used on both encoder (for the MT auxiliary task) and decoder sides. We set the batch size dynamically depending on the input sequence size such that the average batch size is 24 sentences. We use Adam \cite{Kingma2014} with initial learning rate of 0.0005, decayed by 0.5 when the validation BLEU score did not improve over 10 check points initially and 5 check points after the first decay. We initialize attention-passing models using weights from a basic two-stage model trained on the same data.

Following \newcite{Weiss2017}, we lowercase texts and remove punctuation. As speech features, we use 40-dimensional Mel filter bank features with per-speaker mean and variance normalization. We exclude a small number of utterances longer than 1500 frames from training to avoid running out of memory. The encoder-decoder attention is MLP-based, and the decoder uses a single LSTM layer.\footnote{\newcite{Weiss2017} report improvements from deeper decoders, but we encountered stability issues and therefore restricted the decoder to a single layer.} Source text encoders for the multi-task direct model and the cascaded models use two LSTM layers. The number of hidden units is 128 for the encoder-decoder attention MLP, 64 for target character embeddings, 256 for the encoder LSTMs in each direction, and 512 elsewhere. The model uses variational recurrent dropout with probability 0.3 and target character dropout with probability 0.1 \cite{Gal2016}. We apply label smoothing \cite{Szegedy2016} and fix the target embedding norm to 1 \cite{Nguyen2017}. We use beam search with beam size 15 and polynomial length normalization with exponent $1.5$.\footnote{For two-stage and attention-passing models, we apply beam search only for the second stage decoder. We do not employ the two-phase beam search of \newcite{Tu2017} because of its prohibitive memory requirements.}

All BLEU scores are computed on Fisher/Test against 4 references.

\subsection{Cascaded vs.\ Direct Models}
\label{sec:e2e-casc-vs-direct}
We first wish to shed light on the question of whether cascaded or direct models can be expected to perform better. This question has been investigated previously \cite{Weiss2017,Kano2017,Berard2018,Anastasopoulos2018}, but with contradictory findings. We hypothesize that the increased complexity of the direct mapping from speech to translation increases the data requirements of such models. Table~\ref{tab:cascade-vs-direct} compares the direct multi-task model (\S\ref{sec:e2e-direct-multi-task}) against a cascaded model with identical architecture to the respective ASR and MT sub-models of the multi-task model. The direct model is trained with multi-task training on the auxiliary ASR, MT, and AE tasks on the same data which outperformed single-task training considerably in preliminary experiments. As can be seen, the direct model outperforms the traditional cascaded set-up only when both are trained on the full data, but not when using only parts of the training data. This provides evidence in favor of our hypothesis and indicates that direct end-to-end models should be expected to perform strongly only in a case where enough training data is available.

\begin{table}[tb]
\centering
\begin{tabular}{@{}ccc@{}}
\toprule
Training sents.                   & Cascade        & Direct model   \\ 
\midrule
139k                                   & 32.45             & \textbf{35.30} \\ 
69k                                     & \textbf{26.52} & 24.68             \\
35k                                     & \textbf{16.84} & 14.91             \\
14k                                     & \textbf{6.59}   & 6.08               \\
\bottomrule
\end{tabular}
\caption{BLEU scores (4 references) on Fisher/Test for various amounts of training data. The direct (multi-task) model performs best in the full data condition, but the cascaded model is best in all reduced conditions.}
\label{tab:cascade-vs-direct}
\end{table}

\subsection{Two-Stage Models}

Next, we investigate the performance of the two-stage models, for both the basic variant (\S\ref{sec:e2e-multistage-multitask}) and our proposed attention-passing model (\S\ref{sec:e2e-modified-model}). Again, all models are trained in a multi-task fashion by including auxiliary ASR and MT tasks based on the same data. Table~\ref{tab:two-stage} shows the results. The basic two-stage model performs in between the direct and cascaded models from \S\ref{sec:e2e-casc-vs-direct}. \texttt{APM}, the attention-passing model of \S\ref{sec:e2e-modified-model-basic} which is designed to circumvent the negative effects of error propagation, outperforms the basic variant and performs similarly to the direct model. The \texttt{APM} extensions (\S\ref{sec:e2e-cross-connections}, \S\ref{sec:e2e-add-loss}) further improved the results, with the best model outperforming the direct model by 1.40 BLEU points and the basic two-stage model by 2.34 BLEU points absolute. The last row in the table confirms that the block dropout operation contributed to the gains: removing it led to a drop by 0.66 BLEU points.

\begin{table}[tb]
\centering
\begin{tabular}{@{}lc@{}}
\toprule
Model                  & BLEU \\
\midrule
Cascade             & 32.45               \\
Direct                  & 35.30               \\
\midrule
Basic two-stage        & 34.36               \\
\midrule
\texttt{APM}     & 35.31               \\
\texttt{APM} + cross connections    & 36.51               \\
\texttt{APM} + cross conn.\ + additional loss      & \textbf{36.70}               \\
\midrule
Best \texttt{APM} w/o block dropout & 36.04              \\
\bottomrule
\end{tabular}
\caption{Results for cascaded and multi-task models under full training data conditions.}
\label{tab:two-stage}
\end{table}

\subsection{Data Efficiency: Direct Model}
\label{sec:e2e-data-efficiency-direct}

Having established results in favor of our proposed model on the full data, we now examine the data efficiency of the different models. Our experimental strategy is to compare model performance (1) when trained on the full data, (2) when trained on partial data, and (3) when trained on partial speech-to-translation data but full auxiliary (ASR+MT) data.\footnote{An alternative experimental strategy is to train on the full data and then add auxiliary data from other domains to the training. We pursue this strategy in \S\ref{sec:e2e-open-sub} as a more realistic scenario, but point out several problems that lead us to not use this as our main approach: Adding external auxiliary data (1) leads to side-effects due to domain mismatch and (2) severely limits the number of experiments that we can conduct due to the considerably increased training time.}

\begin{figure}[tb]
\includegraphics[width=7.5cm]{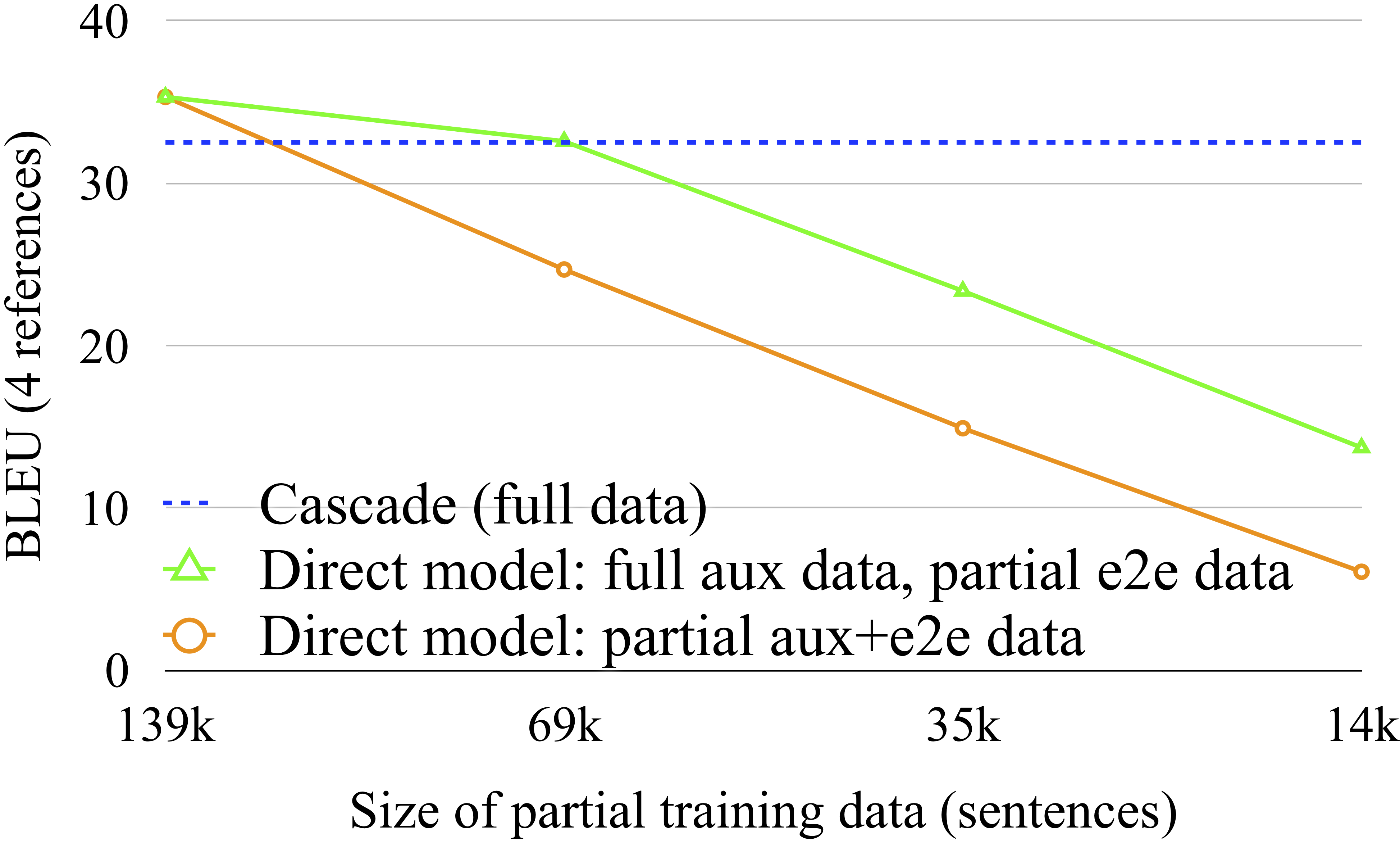}
\caption{Data efficiency for direct (multi-task) model, compared against cascade on full auxiliary data.}
\label{fig:e2e-results-direct-multitask}
\end{figure}

Figure~\ref{fig:e2e-results-direct-multitask} shows the results, comparing the cascaded model against the direct model trained under conditions (1), (2), and (3).\footnote{Note that the above hyper-parameters were selected for best full-data performance and are not re-tuned here.}
Unsurprisingly, the performance of the direct model trained on partial data declines sharply as the amount of data is reduced. Adding auxiliary data through multi-task training improves performance in all cases.
For instance, in the case of 69k speech-to-translation instances, adding the full auxiliary data helps to reach the accuracy of the cascaded model. However, this is already somewhat disappointing because the end-to-end data, which is not available to the cascaded model, no longer yields an advantage. Moreover, reducing the end-to-end data further reveals that multi-task training is not able to close the gap to the cascade. In the scenario with 35k end-to-end instances and full auxiliary data, the direct model underperforms the cascade by 9.14 BLEU points (32.50 vs.\ 23.36), despite being trained on {\it more} data. The unsatisfactory data efficiency in this controlled ablation study strongly indicates that the direct model will also fall behind a cascade that is trained on large amounts of external data. This claim is verified in \S\ref{sec:e2e-open-sub}.

\subsection{Data Efficiency: Two-Stage Models}
\label{sec:e2e-data-efficiency-2stage}

\begin{figure}[tb]
\includegraphics[width=7.5cm]{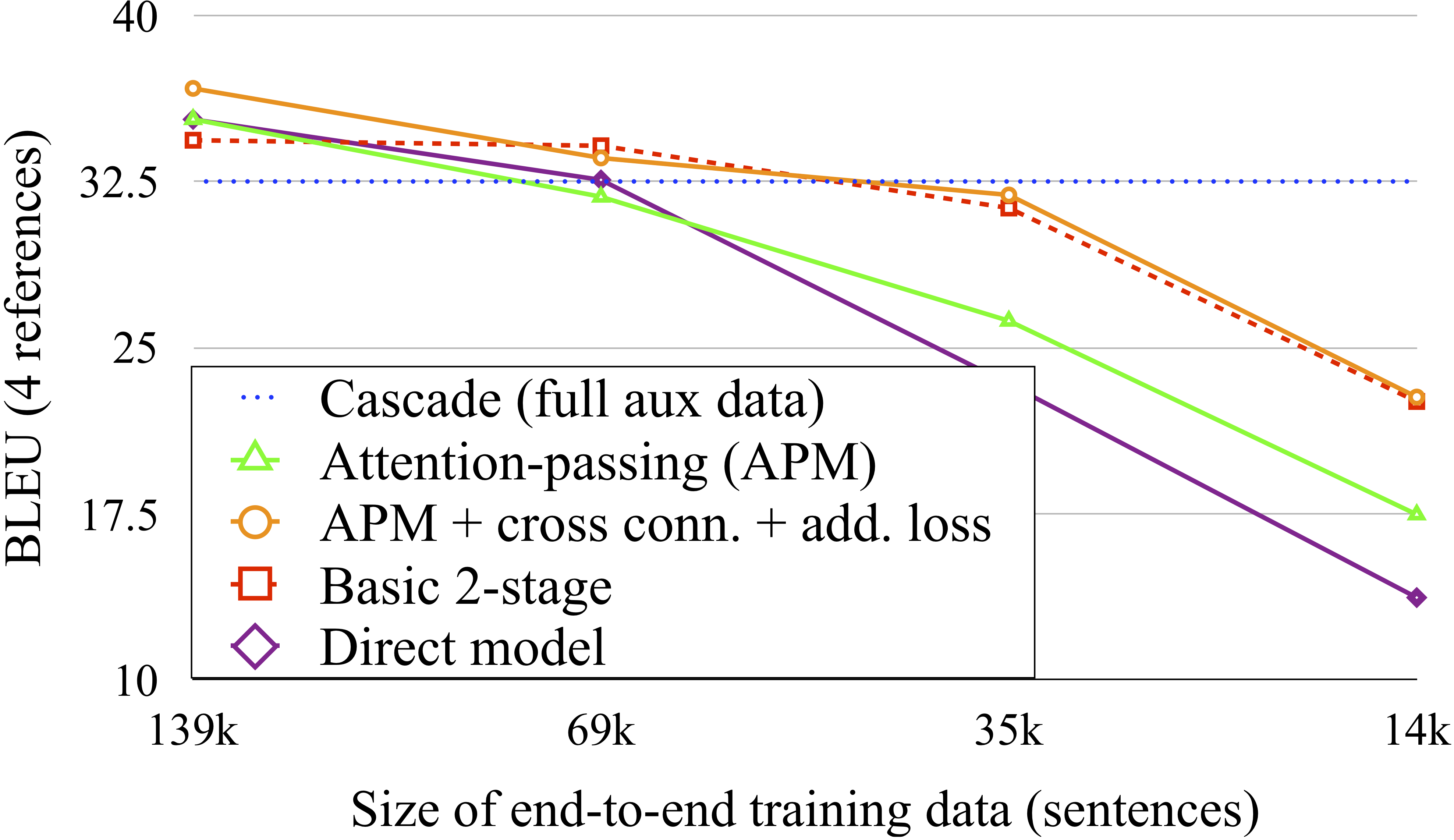}
\caption{Data efficiency across model types. All models use full auxiliary data through multi-task training.}
\label{fig:e2e-results-data-efficiency}
\end{figure}

We showed that the direct model is poor at integrating auxiliary data and heavily depends on sufficient amounts of end-to-end training data. How do two-stage models behave with regards to this data efficiency issue? Figure~\ref{fig:e2e-results-data-efficiency} shows that both the basic two-stage model and the best \texttt{APM} perform reasonably well even when having seen much less end-to-end data. We can explain this by noticing that these models can be naturally decomposed into an ASR sub-model and an MT sub-model, while the direct model needs to add auxiliary sub-models to support multi-task training. Interestingly, the attention-passing model without cross-connections does better than the direct model with regards to data efficiency, but falls behind the basic and best proposed two-stage models. This indicates that access to ASR labels in some form contributes to favorable data efficiency of speech translation models.

\subsection{Adding External Data}
\label{sec:e2e-open-sub}

Our approach for evaluating data efficiency so far has been to assume that end-to-end data is available for only a subset of the available auxiliary data. In practice, we can often train ASR and MT tasks on abundant external data. We therefore run experiments in which we use the full Fisher training data for all tasks as before, and add OpenSubtitle\footnote{\url{http://www.opensubtitles.org/}} data for the auxiliary MT task. We clean and normalize the Spanish-English OpenSubtitle 2018 data \cite{Lison2016} to be consistent with the employed Fisher training data by lowercasing and removing punctuation. We apply a basic length filter and obtain 61 million sentences. During training, we include the same number of sentences from in-domain and out-of-domain MT tasks in each minibatch in order to prevent degradation due to domain mismatch. Our models converged before a full pass over the OpenSubtitle data, but needed between two and three times more steps than the in-domain model to converge.

Table~\ref{tab:opensub} shows that all models were able to benefit from the added data. However, when examining the relative gains we can see that both the cascaded model and the models with two attention stages benefitted about twice as much from the external data as the direct model. In fact, the basic two-stage model now slightly surpasses the direct model, and the best \texttt{APM} is ahead of the basic two-stage model by almost the same absolute difference as before (2.36 BLEU points). The superior relative gains show that our findings from \S\ref{sec:e2e-data-efficiency-direct} and \S\ref{sec:e2e-data-efficiency-2stage}, namely that two-stage models are much more efficient at exploiting auxiliary training data, generalizes to the setting in which large amounts of out-of-domain data are added to the MT task. Out-of-domain data is often much easier to obtain, and we can therefore conclude that the proposed approach is preferable in many practically relevant situations. Because these experiments are very expensive to conduct, we leave experiments with external ASR data for future work.

\begin{table}[tb]
\begin{tabular}{lcc}
\toprule
Model                      & Fisher & Fisher+OpenSub \\ 
\hline
Cascade                  & 32.45  &   34.58  (+6.2\% rel.)              \\
Direct model            & 35.30  &   36.45  (+3.2\% rel.)               \\
Basic two-stage       & 34.36  &   36.91  (+6.9\% rel.)               \\
Best \texttt{APM}     & 36.70  &  38.81 (+5.4\% rel.)                          \\
\bottomrule
\end{tabular}
\caption{Adding auxiliary OpenSubtitles MT data to the training. The two-stage models benefit much more strongly than the direct model, with our proposed model yielding the strongest overall results.}
\label{tab:opensub}
\end{table}

\subsection{Error Propagation}
\label{sec:analysis-error-prop}
To better understand the impact of error propagation, we analyze how improved or degraded ASR labels impact the translation results. This experiment is applicable to \texttt{APM}, two-stage model and the cascade, but not to the direct model which does not compute intermediate ASR outputs. We analyze three different settings: using the standard decoded ASR labels, replacing these labels with the gold labels, or artificially degrading the decoded labels by randomly introducing 10\% of substitution, insertion, and deletion noise \cite{Sperber2017a}. Intuitively, models that suffer from error propagation issues are expected to rely most heavily on these intermediate labels and would therefore be most impacted by both degraded and improved labels.

Table~\ref{tab:error-prop} shows the results. Unsurprisingly, the cascade responds most strongly to improved or degraded noise, confirming that it is severely impacted by error propagation. The \texttt{APM}, which does not directly expose the labels to the translation sub-model, is much less impacted. However, the impact is still more significant than perhaps expected, suggesting that improved attention models that are more robust to decoding errors \cite{Chorowski2015,Tjandra2017} may serve to further improve our model in the future. Note that the \texttt{APM} benefits poorly from gold ASR labels, which is expected because gold labels only improve the ASR alignments and by extension the passed context vectors, but these are quite robust against decoding errors in the first place.

The basic two-stage model is impacted significantly, although less strongly than the cascade, in line with our claim that such models are subject to error propagation despite being end-to-end trainable. Note that it falls behind the cascade for gold labels, despite both models being seemingly identical under this condition. This can be explained by the cascaded model's use of beam search and greater number of encoder layers.

Somewhat contrary to our expectations, \texttt{APM} with cross connections appears equally subject to error propagation despite the block dropout on these connections, displaying the same accuracy gains across the three different settings. This suggests future explorations toward model variants with an even better trade-off between overall accuracy, data efficiency, and amount of degradation due to error propagation.

\begin{table}[tb]
\begin{tabular}{@{}lccc@{}}
\toprule
Labels                  & Gold                  & Decod.      & Perturbed \\ \midrule
Cascade               & 58.15 (+44\%)   & 32.45            &   25.67 (-26\%)      \\
B2S                 & 56.60 (+39\%)   & 34.36            &   28.81 (-19\%)     \\
\texttt{APM}          & 40.70 (+13\%)   & 35.31            &   31.96  (-10\%)    \\
+ cross                  & 58.29 (+37\%)   & 36.70            &  30.48  (-20\%)     \\
\bottomrule
\end{tabular}
\caption{Effect of altering the ASR labels for different models  as a measure for robustness against error propagation. We compare results for the cascade, the basic two-stage model (B2S), and \texttt{APM} without and with cross connections. Percentages are relative to the results for unaltered (decoded) ASR labels.}
\label{tab:error-prop}
\end{table}

\subsection{Robustness of ASR Attentions}
\label{sec:e2e-robust_attentions}

\begin{figure}[tb]
\centering
\includegraphics[width=6.7cm]{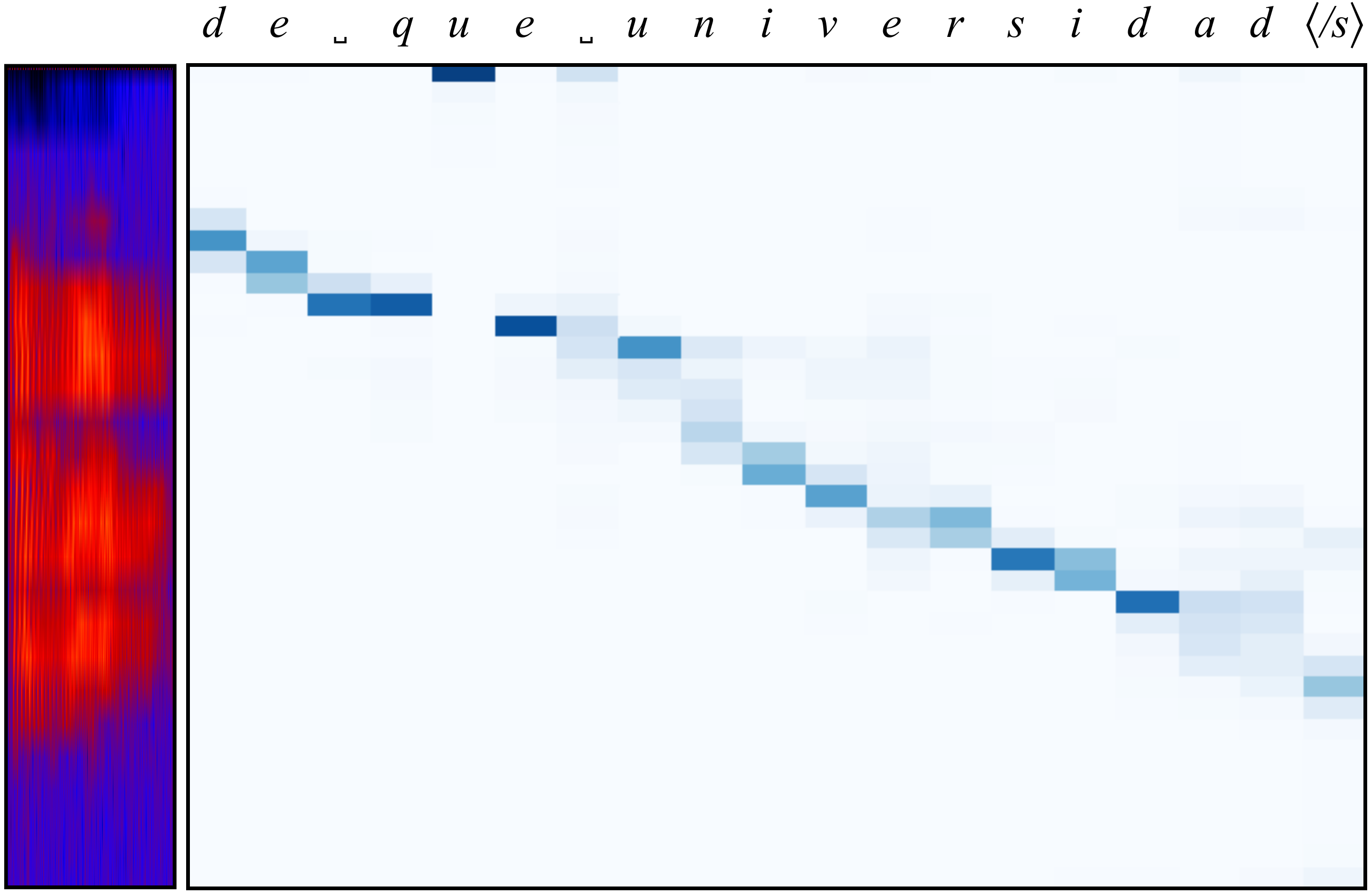}
\caption{ASR attentions when force-decoding the oracle transcripts.}
\label{fig:att-oracle}
\includegraphics[width=6.7cm]{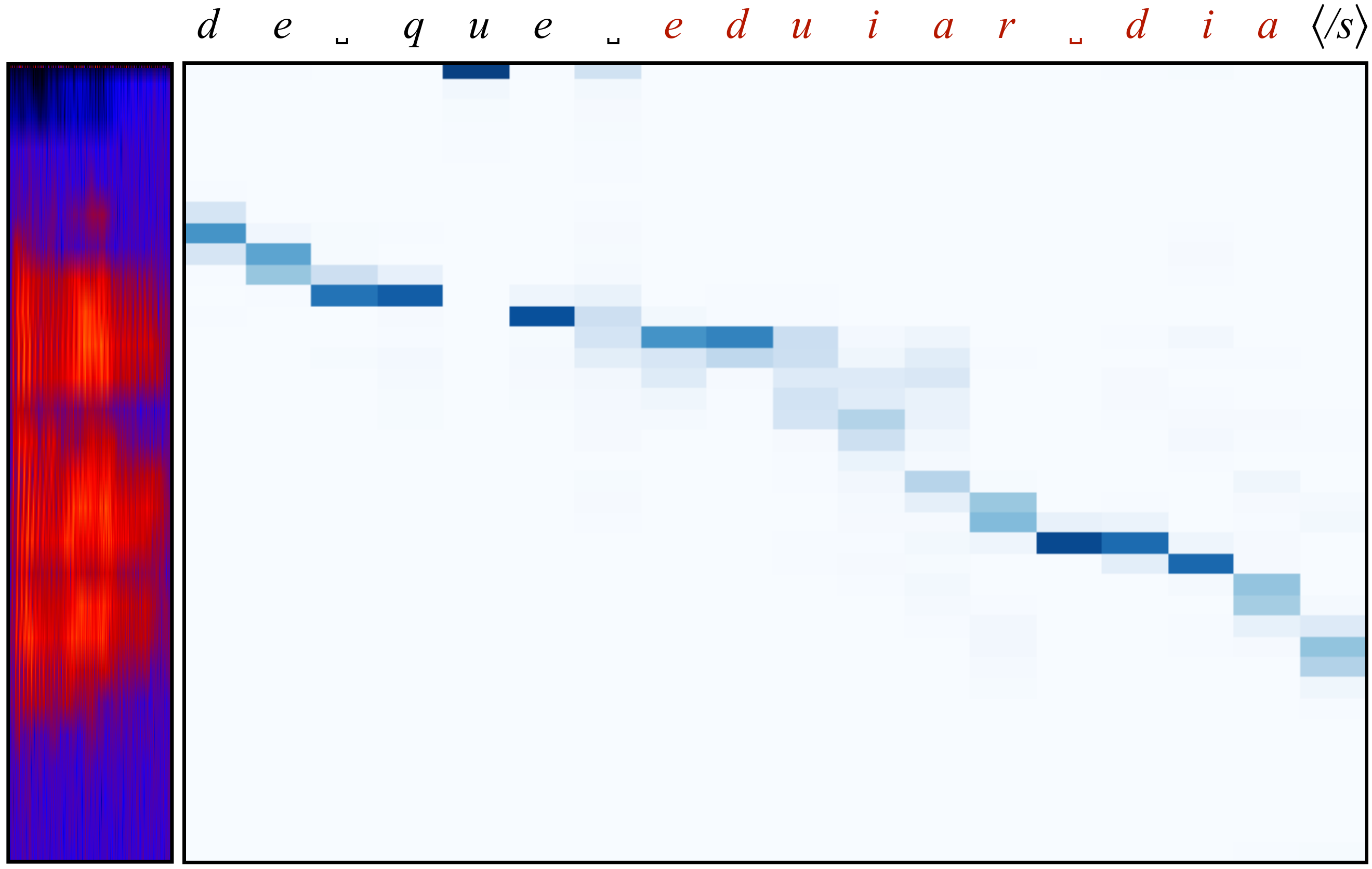}
\caption{ASR attentions after regular decoding.}
\label{fig:att-dec}
\end{figure}

The attention-passing model was motivated by the assumption that attention scores are relatively robust against recognition errors. We perform a qualitative analysis to validate this assumption. Figure~\ref{fig:att-oracle} shows the first-stage attention matrix when force-decoding the reference transcript, while Figure~\ref{fig:att-dec} shows the same for regular decoding, which for this utterance produced significant errors. Despite the errors, we can see that the attention matrices are very similar. We manually inspected the first 100 test attention matrices and confirm that this behavior occurs very consistently. Further quantitative evidence is given in \S\ref{sec:analysis-error-prop} which showed that the attention-passing model is more resistent to error propagation than the other models.

\section{Prior Work}
Model architectures similar to what we have referred to as the basic two-stage model have first been used by \newcite{Tu2017} for a reconstruction task, where the first stage performs translation and the second stage attempts to reconstruct the original inputs based on the outputs of the first stage. A second variant of a similar architecture are \newcite{Xia2017}'s deliberation networks, where the second stage refines or polishes the outputs of the first stage. For our purposes, the first stage performs speech recognition, a natural intermediate representation for the speech translation task, corresponding to the second stage output. \newcite{Toshniwal2017} explore a different way of lower-level supervision during training of an attentional speech recognizer by jointly training an auxiliary phoneme recognizer based on a lower layer in the acoustic encoder. Similarly to the discussed multi-task direct model, this approach discards many of the learned parameters when used on the main task and consequently may also suffer from data efficiency issues.

Direct end-to-end speech translation models were first used by \cite{Duong2016}, although the authors did not actually evaluate translation performance. \newcite{Weiss2017} extended this model into a multi-task model and report excellent translation results. Our baselines do not match their results, despite considerable efforts. We note that other research groups have encountered similar replicability issues \cite{Bansal2018}, explanations include the lack of a large GPU cluster to perform ASGD training, as well as to explore an ideal number of training schedules and other hyper-parameter settings. \newcite{Berard2018} explored the translation of audio books with direct models and report reasonable results, but do not outperform a cascaded baseline. \newcite{Kano2017} have first used a basic two-stage model for speech translation. They use a pretraining strategy for the individual sub-models, related to our multi-task approach, but do not attempt to integrate auxiliary data. Moreover, the authors only evaluated the translation of synthesized speech, which greatly simplifies training and may not lead to generalizable conclusions, as indicated by the fact that they were actually able to outperform a translation model that used the gold transcripts as input. \newcite{Anastasopoulos2018} conduct experiments on low-resource speech translation and employ a triangle model that can be seen as a combination of a direct model and a two-stage model, but is not easily trainable in a multi-task fashion. It is therefore not a suitable choice for exploiting auxiliary data in order to compete with cascaded models under well-resourced data conditions. Finally, contemporaneous work explores transferring knowledge from high-resource to low-resource languages \cite{Bansal2018a}.

\section{Conclusion}
This work explored {\it direct} and {\it two-stage} models for speech translation with the aim of obtaining models that are strong not only in favorable data conditions, but are also efficient at exploiting auxiliary data. We started by demonstrating that direct models do outperform cascaded models, but only when enough data is available, shedding light on inconclusive results from prior work. We further showed that these models are poor at exploiting auxiliary data, making them a poor choice in realistic situations. We motivated the use of two-stage models by their ability to overcome this shortcoming of the direct models, and found that two-stage models are in fact more data-efficient, but suffer from error propagation issues. We addressed this by introducing a novel attention-passing model that alleviates error propagation issues, as well as several model variants. The best proposed model outperforms all other tested models and is much more data efficient than the direct model, allowing this model to compete with cascaded models even under realistic assumptions with auxiliary data available. Analysis showed that there seems to be a trade-off between data efficiency and error propagation. Avenues for future work include testing better ASR attention models, adding other types of external data such as ASR data, unlabeled speech, or monolingual texts, and exploring further model variants.

\section*{Acknowledgments}
We thank Adam Lopez, Stefan Constantin and the anonymous reviewers for their helpful comments. The work leading to these results has received funding from the European Union under grant agreement no 825460.

\bibliography{library.bib}

\begin{thebibliography}{28}
\expandafter\ifx\csname natexlab\endcsname\relax\def\natexlab#1{#1}\fi

\bibitem[{Ammar et~al.(2016)Ammar, Mulcaire, Ballesteros, Dyer, and
  Smith}]{Ammar2016}
Waleed Ammar, George Mulcaire, Miguel Ballesteros, Chris Dyer, and Noah~A.
  Smith. 2016.
\newblock \href {http://arxiv.org/abs/1602.01595} {{Many Languages, One
  Parser}}.
\newblock \emph{Transactions of the Association for Computational Linguistics
  (TACL)}, 4:431--444,.

\bibitem[{Anastasopoulos and Chiang(2018)}]{Anastasopoulos2018}
Antonios Anastasopoulos and David Chiang. 2018.
\newblock \href {http://arxiv.org/abs/1802.06655} {{Tied Multitask Learning for
  Neural Speech Translation}}.
\newblock In \emph{North American Chapter of the Association for Computational
  Linguistics (NAACL)}, New Orleans, USA.

\bibitem[{Bahdanau et~al.(2015)Bahdanau, Cho, and Bengio}]{Bahdanau2014}
Dzmitry Bahdanau, KyungHyun Cho, and Yoshua Bengio. 2015.
\newblock \href {http://arxiv.org/abs/1409.0473} {{Neural Machine Translation
  by Jointly Learning to Align and Translate}}.
\newblock In \emph{International Conference on Representation Learning (ICLR)},
  San Diego, USA.

\bibitem[{Bansal et~al.(2018)Bansal, Kamper, Livescu, Lopez, and
  Goldwater}]{Bansal2018}
Sameer Bansal, Herman Kamper, Karen Livescu, Adam Lopez, and Sharon Goldwater.
  2018.
\newblock \href {http://arxiv.org/abs/1803.09164} {{Low-Resource Speech-to-Text
  Translation}}.
\newblock In \emph{Annual Conference of the International Speech Communication
  Association (InterSpeech)}.

\bibitem[{Bansal et~al.(2019)Bansal, Kamper, Livescu, Lopez, and
  Goldwater}]{Bansal2018a}
Sameer Bansal, Herman Kamper, Karen Livescu, Adam Lopez, and Sharon Goldwater.
  2019.
\newblock \href {http://arxiv.org/abs/1809.01431} {{Pre-training on
  high-resource speech recognition improves low-resource speech-to-text
  translation}}.
\newblock In \emph{North American Chapter of the Association for Computational
  Linguistics (NAACL)}.

\bibitem[{B{\'{e}}rard et~al.(2018)B{\'{e}}rard, Besacier, Kocabiyikoglu, and
  Pietquin}]{Berard2018}
Alexandre B{\'{e}}rard, Laurent Besacier, Ali~Can Kocabiyikoglu, and Olivier
  Pietquin. 2018.
\newblock \href {http://arxiv.org/abs/1802.04200} {{End-to-End Automatic Speech
  Translation of Audiobooks}}.
\newblock In \emph{International Conference on Acoustics, Speech and Signal
  Processing (ICASSP)}, Calgary, Canada.

\bibitem[{Chan et~al.(2016)Chan, Jaitly, Le, and Vinyals}]{Chan2016}
William Chan, Navdeep Jaitly, Quoc~V. Le, and Oriol Vinyals. 2016.
\newblock \href {https://doi.org/10.1109/72.279181} {{Listen, attend and spell:
  A neural network for large vocabulary conversational speech recognition}}.
\newblock In \emph{Acoustics, Speech and Signal Processing (ICASSP)}.

\bibitem[{Chorowski et~al.(2015)Chorowski, Bahdanau, Serdyuk, Cho, and
  Bengio}]{Chorowski2015}
Jan~K. Chorowski, Dzmitry Bahdanau, Dmitriy Serdyuk, Kyunghyun Cho, and Yoshua
  Bengio. 2015.
\newblock \href
  {http://papers.nips.cc/paper/5847-attention-based-models-for-speech-recognition}
  {{Attention-Based Models for Speech Recognition}}.
\newblock In \emph{Advances in Neural Information Processing Systems (NIPS)},
  pages 577--585.

\bibitem[{Cieri et~al.(2004)Cieri, Miller, and Walker}]{Cieri2004}
Christopher Cieri, David Miller, and Kevin Walker. 2004.
\newblock \href
  {https://pdfs.semanticscholar.org/a723/97679079439b075de815553c7b687ccfa886.pdf}
  {{The Fisher Corpus: a Resource for the Next Generations of Speech-to-Text.}}
\newblock In \emph{Language Resources and Evaluation (LREC)}, pages 69--71.

\bibitem[{Duong et~al.(2016)Duong, Anastasopoulos, Chiang, Bird, and
  Cohn}]{Duong2016}
Long Duong, Antonios Anastasopoulos, David Chiang, Steven Bird, and Trevor
  Cohn. 2016.
\newblock \href {http://www.aclweb.org/anthology/N16-1109} {{An Attentional
  Model for Speech Translation Without Transcription}}.
\newblock In \emph{North American Chapter of the Association for Computational
  Linguistics (NAACL)}, pages 949--959, San Diego, USA.

\bibitem[{Gal and Ghahramani(2016)}]{Gal2016}
Yarin Gal and Zoubin Ghahramani. 2016.
\newblock \href
  {http://papers.nips.cc/paper/6241-a-theoretically-grounded-application-of-dropout-in-recurrent-neural-networks.pdf}
  {{A Theoretically Grounded Application of Dropout in Recurrent Neural
  Networks}}.
\newblock In \emph{Neural Information Processing Systems Conference (NIPS)},
  pages 1019--1027, Barcelona, Spain.

\bibitem[{Kano et~al.(2017)Kano, Sakti, and Nakamura}]{Kano2017}
Takatomo Kano, Sakriani Sakti, and Satoshi Nakamura. 2017.
\newblock \href
  {https://www.isca-speech.org/archive/Interspeech{\_}2017/pdfs/0944.PDF}
  {{Structured-based Curriculum Learning for End-to-end English-Japanese Speech
  Translation}}.
\newblock In \emph{Annual Conference of the International Speech Communication
  Association (InterSpeech)}, pages 2630--2634.

\bibitem[{Kingma and Ba(2014)}]{Kingma2014}
Diederik~P. Kingma and Jimmy~L. Ba. 2014.
\newblock \href {https://arxiv.org/pdf/1412.6980.pdf} {{Adam: A Method for
  Stochastic Optimization}}.
\newblock In \emph{International Conference on Learning Representations
  (ICLR)}, Banff, Canada.

\bibitem[{Kocabiyikoglu et~al.(2018)Kocabiyikoglu, Besacier, and
  Kraif}]{Kocabiyikoglu2018}
Ali~Can Kocabiyikoglu, Laurent Besacier, and Olivier Kraif. 2018.
\newblock \href {https://arxiv.org/abs/1802.03142} {{Augmenting Librispeech
  with French Translations: A Multimodal Corpus for Direct Speech Translation
  Evaluation}}.
\newblock In \emph{Language Resources and Evaluation (LREC)}, Miyazaki, Japan.

\bibitem[{Lison and Tiedemann(2016)}]{Lison2016}
Pierre Lison and J{\"{o}}rg Tiedemann. 2016.
\newblock \href {http://stp.lingfil.uu.se/{~}joerg/paper/opensubs2016.pdf}
  {{OpenSubtitles2016: Extracting Large Parallel Corpora from Movie and TV
  Subtitles}}.
\newblock In \emph{Conference on Language Resources and Evaluation (LREC)},
  pages 923--929, Portoro{\v{z}}, Slovenia.

\bibitem[{Neubig et~al.(2018)Neubig, Sperber, Wang, Felix, Matthews,
  Padmanabhan, Qi, Sachan, Arthur, Godard, Hewitt, Riad, and
  Wang}]{neubig18xnmt}
Graham Neubig, Matthias Sperber, Xinyi Wang, Matthieu Felix, Austin Matthews,
  Sarguna Padmanabhan, Ye~Qi, Devendra~Singh Sachan, Philip Arthur, Pierre
  Godard, John Hewitt, Rachid Riad, and Liming Wang. 2018.
\newblock \href {https://arxiv.org/pdf/1803.00188.pdf} {{XNMT: The eXtensible
  Neural Machine Translation Toolkit}}.
\newblock In \emph{Conference of the Association for Machine Translation in the
  Americas (AMTA) Open Source Software Showcase}, Boston, USA.

\bibitem[{Nguyen and Chiang(2018)}]{Nguyen2017}
Toan~Q. Nguyen and David Chiang. 2018.
\newblock \href {https://arxiv.org/pdf/1710.01329.pdf} {{Improving Lexical
  Choice in Neural Machine Translation}}.
\newblock In \emph{North American Chapter of the Association for Computational
  Linguistics (NAACL)}, New Orleans, USA.

\bibitem[{Niehues and Cho(2017)}]{Niehues2017}
Jan Niehues and Eunah Cho. 2017.
\newblock \href {https://arxiv.org/pdf/1708.00993.pdf} {{Exploiting Linguistic
  Resources for Neural Machine Translation Using Multi-task Learning}}.
\newblock In \emph{Conference on Machine Translation (WMT)}, pages 80--89,
  Copenhagen, Denmark.

\bibitem[{Panayotov et~al.(2015)Panayotov, Chen, Povey, and
  Khudanpur}]{Panayotov2015}
Vassil Panayotov, Guoguo Chen, Daniel Povey, and Sanjeev Khudanpur. 2015.
\newblock \href
  {http://www.danielpovey.com/files/2015{\_}icassp{\_}librispeech.pdf}
  {{Librispeech: an ASR corpus based on public domain audio books}}.
\newblock In \emph{Acoustics, Speech and Signal Processing (ICASSP)}, pages
  5206--5210, Brisbane, Australia.

\bibitem[{Post et~al.(2013)Post, Kumar, Lopez, Karakos, Callison-Burch, and
  Khudanpur}]{Post2013}
Matt Post, Gaurav Kumar, Adam Lopez, Damianos Karakos, Chris Callison-Burch,
  and Sanjeev Khudanpur. 2013.
\newblock \href {http://cs.jhu.edu/{~}gkumar/papers/post2013improved.pdf}
  {{Improved Speech-to-Text Translation with the Fisher and Callhome
  Spanish--English Speech Translation Corpus}}.
\newblock In \emph{International Workshop on Spoken Language Translation
  (IWSLT)}, Heidelberg, Germany.

\bibitem[{Sperber et~al.(2017)Sperber, Niehues, and Waibel}]{Sperber2017a}
Matthias Sperber, Jan Niehues, and Alex Waibel. 2017.
\newblock \href {http://isl.anthropomatik.kit.edu/pdf/Sperber2017b.pdf}
  {{Toward Robust Neural Machine Translation for Noisy Input Sequences}}.
\newblock In \emph{International Workshop on Spoken Language Translation
  (IWSLT)}, Tokyo, Japan.

\bibitem[{Szegedy et~al.(2016)Szegedy, Vanhoucke, Ioffe, Shlens, and
  Wojna}]{Szegedy2016}
Christian Szegedy, Vincent Vanhoucke, Sergey Ioffe, Jon Shlens, and Zbigniew
  Wojna. 2016.
\newblock \href
  {https://www.cv-foundation.org/openaccess/content{\_}cvpr{\_}2016/papers/Szegedy{\_}Rethinking{\_}the{\_}Inception{\_}CVPR{\_}2016{\_}paper.pdf}
  {{Rethinking the Inception Architecture for Computer Vision}}.
\newblock In \emph{Computer Vision and Pattern Recognition (CVPR)}, pages
  2818--2826, Las Vegas, USA.

\bibitem[{Tjandra et~al.(2017)Tjandra, Sakti, and Nakamura}]{Tjandra2017}
Andros Tjandra, Sakriani Sakti, and Satoshi Nakamura. 2017.
\newblock \href {http://arxiv.org/abs/1705.08091} {{Local Monotonic Attention
  Mechanism for End-to-End Speech Recognition}}.
\newblock In \emph{International Joint Conference on Natural Language
  Processing (IJCNLP)}, pages 431--440.

\bibitem[{Toshniwal et~al.(2017)Toshniwal, Tang, Lu, and
  Livescu}]{Toshniwal2017}
Shubham Toshniwal, Hao Tang, Liang Lu, and Karen Livescu. 2017.
\newblock \href {http://arxiv.org/abs/1704.01631} {{Multitask Learning with
  Low-Level Auxiliary Tasks for Encoder-Decoder Based Speech Recognition}}.
\newblock In \emph{Annual Conference of the International Speech Communication
  Association (InterSpeech)}, Stockholm, Sweden.

\bibitem[{Tu et~al.(2017)Tu, Liu, Shang, Liu, and Li}]{Tu2017}
Zhaopeng Tu, Yang Liu, Lifeng Shang, Xiaohua Liu, and Hang Li. 2017.
\newblock \href {http://arxiv.org/abs/arXiv:1611.01874v2} {{Neural Machine
  Translation with Reconstruction}}.
\newblock In \emph{Conference on Artificial Intelligence (AAAI)}.

\bibitem[{Weiss et~al.(2017)Weiss, Chorowski, Jaitly, Wu, and Chen}]{Weiss2017}
Ron~J. Weiss, Jan Chorowski, Navdeep Jaitly, Yonghui Wu, and Zhifeng Chen.
  2017.
\newblock \href {http://arxiv.org/abs/1703.08581} {{Sequence-to-Sequence Models
  Can Directly Transcribe Foreign Speech}}.
\newblock In \emph{Annual Conference of the International Speech Communication
  Association (InterSpeech)}, Stockholm, Sweden.

\bibitem[{Xia et~al.(2017)Xia, Tian, Wu, Lin, Qin, Yu, and Liu}]{Xia2017}
Yingce Xia, Fei Tian, Lijun Wu, Jianxin Lin, Tao Qin, Nenghai Yu, and Tie-Yan
  Liu. 2017.
\newblock \href {https://doi.org/10.1016/S0960-1481(96)00129-2} {{Deliberation
  Networks: Sequence Generation}}.
\newblock In \emph{Neural Information Processing Systems Conference (NIPS)},
  Long Beach, USA.

\bibitem[{Zhang et~al.(2017)Zhang, Chan, and Jaitly}]{Zhang2017e}
Yu~Zhang, William Chan, and Navdeep Jaitly. 2017.
\newblock \href {http://arxiv.org/abs/1610.03022} {{Very Deep Convolutional
  Networks for End-to-End Speech Recognition}}.
\newblock In \emph{International Conference on Acoustics, Speech and Signal
  Processing (ICASSP)}.

\end{thebibliography}
\bibliographystyle{acl_natbib}

\end{document}